# QuickEdit: Editing Text & Translations by Crossing Words Out

**David Grangier**   **Michael Auli**

Facebook AI Research
Menlo Park, CA

## Abstract

We propose a framework for computer-assisted text editing. It applies to translation post-editing and to paraphrasing. Our proposal relies on very simple interactions: a human editor modifies a sentence by marking tokens they would like the system to change. Our model then generates a new sentence which reformulates the initial sentence by avoiding marked words. The approach builds upon neural sequence-to-sequence modeling and introduces a neural network which takes as input a sentence along with change markers. Our model is trained on translation bitext by simulating post-edits. We demonstrate the advantage of our approach for translation post-editing through simulated post-edits. We also evaluate our model for paraphrasing through a user study.

## 1 Introduction

Computers can help humans edit text more efficiently. In particular, statistical models are used for that purpose, for instance to help correct spelling mistakes (Brill and Moore, 2000) or suggest likely completions of a sentence (Bickel et al., 2005). In this work, we rely on statistical learning to enable a computer to rephrase a sentence by only pointing at words that should be avoided. Specifically, we consider the task of reformulating either a sentence, i.e. *paraphrasing* (Quirk et al., 2004), or a translation, i.e. *translation post-editing* (Koehn, 2009b). Paraphrasing reformulates a sentence with different words preserving its meaning, while translation post-editing takes a candidate translation along with the corresponding source sentence and improves it.

Our proposal relies on very simple interactions: a human editor modifies a sentence by selecting tokens they would like the system to replace and no other feedback. Our system then generates a new sentence which reformulates the initial sentence by avoiding the word types from the selected tokens. Our approach builds upon neural sequence-to-sequence and introduces a neural network which takes as input a sentence along with token markers. We introduce a novel attention-based architecture suited to this goal and propose a training procedure based on simulated post-edits on translation bitext (§3). This approach allows to get substantial modifications of the initial sentence – including deletion, reordering and insertion of multiple words – with limited user effort.

Our experiments (§4) relies on large scale simulated post-edits. They show that our model outperforms our post-editing baseline by up to 5 BLEU points on WMT'14 English-German and WMT'14 German-English translation. The advantage of our method is also highlighted in monolingual settings, where we analyze the quality of the paraphrases generated by our model in a user study.

Before introducing our method (§3) and its empirical evaluation (§4), we describe related work in the next section.

## 2 Related Work

Our work builds upon previous research on neural machine translation, machine translation post-editing, and computer-assisted editing.

### 2.1 Neural Machine Translation

Statistical machine translation systems models automatically translate text relying on large corpora of bitext, i.e. corresponding pairs of sentences in the source and target language (Koehn, 2009a). Recently, machine translation systems based on neural networks have emerged as an effective approach to this problem (Sutskever et al., 2014). Neural networks are a departure from count-based translation systems, e.g. phrase-based systems, which used to dominate the field (Koehn, 2009a).

Research in Neural Machine Translation (NMT) focuses notably on identifying appropri-

ate neural architecture. Cho et al. (2014) and Suskever et al. (2014) proposed encoder/decoder models. These models consist of a Recurrent Neural Network (RNN) mapping the source sentence sentence into a latent vector (encoder). This vector conditions an RNN language model (decoder) which generates the target sentence (Mikolov et al., 2010; Graves, 2013). Bahdanau et al. (2014) adds attention to these models, which leverages that the explanation for a given target word in generally localized around a few source words. Recently, new architectures have proposed to replace recurrent modules with convolutions (Gehring et al., 2017) or self-attention (Vaswani et al., 2017) to further increase accuracy. These architecture also perform attention at more than one decoder layer, allowing for more complex attention patterns. In this work, we build upon the architecture of Gehring et al. (2017) since this model offers a good trade-off between high accuracy and fast decoding.

## 2.2 Translation Post-Editing

Post-editing leverages a machine translation system and enable human translators to edit its output with different levels of computer assistance. This enables improving machine translation outputs with lesser effort than purely manual translation.

Green et al. (2014) implement such a system relying on a phrase-based translation system. The system presents an initial translation to the user who can accept a prefix and select among the most likely postfix iteratively. Similar ideas relying on decoding with prefix constrains are common in post-translation (Langlais et al., 2000; Koehn, 2009b; Barrachina et al., 2009). Recently, these approaches based on left-to-right decoding have been extended to neural machine translation (Peris et al., 2017).

Closer to our work, Marie and Max (2015) propose light-weight interactions based on accepting/rejecting spans from the output of a statistical machine translation system. The user labels each span that should appear in the final translation. Unmarked spans are assumed to be undesirable and the system removes any entries that could generate those spans from the phrase table. The phrase table is modified such that only positively marked target spans are allowed to explain the corresponding source phrases.

Compared to their work, we rely on similar interactions but we do not require the user to label every token as either accepted or rejected. The user only needs to mark a few rejections. Also, we build on a more accurate neural translation model which is not amenable to phrase table editing. Finally, our method is equally applicable to the monolingual editing of regular text.

Automatic post-editing (APE) (Lagarda et al., 2009), i.e. a process which automatically modifies an MT output without human guidance (Lagarda et al., 2009), is also an active area of research. Although APE shares similarities to classical post-editing, it is beyond the scope of this paper.

## 2.3 Computer-Assisted Text Editing

Computer assisted text editing has been introduced with interactive computer terminals (Irons and Djorup, 1972). Its first achievement was to simplify the insertion, deletion, and copy of text compared to typewriters. Computers then enabled the emergence of computerized language assistance tools such as spelling correctors (Brill and Moore, 2000) or next word suggestions (Bickel et al., 2005).

More recently, research has focused on generating paraphrases (Bannard and Callison-Burch, 2005; Mallinson et al., 2017), compressing sentences (Rush et al., 2015) or simplifying sentences (Nisioi et al., 2017). This type of work expands the possibilities for *interactive* text generation tools, like our work.

Related to our work, Filippova et al. (2015) considers the task of predicting which tokens can be removed from a sentence without modifying its meaning relying on a recurrent neural network. Our work pursues a different goal since our model does not predict which token to remove, as the user provides this information. Our generation is more involved as our model rephrases the sentences, which includes introducing new words, reordering text, inflecting nouns and verbs, etc. Guu et al. (2017) considers generating text with latent edits. Their goal is not to enable users to control which words need to be changed in an initial sentence but to enable sampling valid English sentences with high lexical overlap around a starting sentence. Contrary to paraphrasing, such samples might introduce negations and other changes impacting meaning.

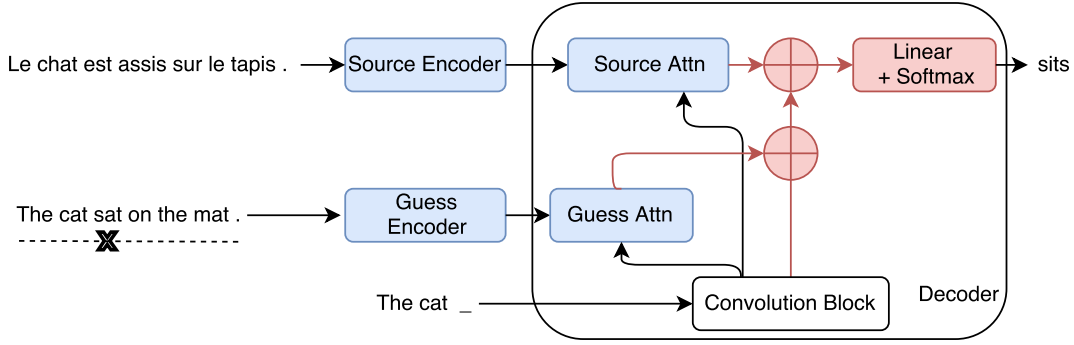

Figure 1: QuickEdit architecture for translation post-editing. The decoder attends to both encodings, one for the source and one for the initial translation (guess) with deletion markers (X on the diagram). Our simplified schema shows one convolutional block and single-hop attention for readability.

## 3 QuickEdit

QuickEdit is our sequence-to-sequence model for post-editing via delete actions. This model takes as input a *source* sentence and an initial *guess* target sentence annotated with change markers. It then aims to improve upon the guess by generating a better target sentence which avoids the marked tokens.

### 3.1 Model Architecture

Our model builds upon the architecture of Gehring et al. (2017). This model is a sequence to sequence neural model with attention. Both the encoder and decoder are deep convolutional networks with residual connections. The model performs *multi-hop* attention, i.e. each layer of the decoder attends to the encoder outputs. Our architecture choice is motivated by the accuracy of this model along with its computational efficiency.

QuickEdit adds a *second* encoder to represent the annotated guess sentence. It also duplicates every attention layer to allow the decoder to attend both to the source and the guess sentences. Dual attention has been introduced recently in the context of automatic post-editing (Novak et al., 2016; Libovický and Helcl, 2017). Our work is however the first work to introduce dual attention in a multi-hop architecture. Figure 1 illustrates our architecture.

The encoder of the initial guess takes as input a target sentence $t$ annotated with binary change labels $c$, i.e.

$$g = \{g_i\}_{i=1}^{l_g} \text{ where } \forall i, g_i = (t_i, c_i)$$

in which $l_g$ denotes the length of the guess, $t_i$ is an index in the target vocabulary and $c_i$ is a binary variable with 1 indicating a request to change the token by the user and 0 indicating no user preference. The first layer of the encoder maps this sequence to two embedding sequences, i.e. a sequence of target word embeddings and a sequence of positional embeddings. Compared to (Gehring et al., 2017), we extend the positional embedding to contain two types of vectors, positional vectors associated with positions $i$ where $c_i = 0$ and positional vectors associated with positions $i$ where $c_i = 1$. Like all parameters in the system, both sets of embeddings are learned to maximize the log-likelihood of the training reference sentences conditioned on the source, annotated guess pairs.

The attention over two sentences is simple. Both source and guess encoders produce a sequence of key and value pairs. We denote the output of the source encoder as $\{(k_i^s, v_i^s)\}_{i=1}^{l_s}$ and the output of the guess encoder as $\{(k_i^g, v_i^g)\}_{i=1}^{l_g}$. At each decoder layer $k$ and time step $j$, the decoder produces a latent state vector $h_j^k$, this vector attends to the output of the source encoder,

$$a_i^s = \exp\left(h_j^k \cdot k_i^s\right) / \sum_l \exp\left(h_j^k \cdot k_l^s\right)$$

and the guess encoder,

$$a_i^g = \exp\left(h_j^k \cdot k_i^g\right) / \sum_l \exp\left(h_j^k \cdot k_l^g\right).$$

This attention weights are used to summarize the values of the source $\sum_i a_i^s v_i^s$ and the guess $\sum_i a_i^s v_i^g$ respectively. The attention module then averages these two vectors $\frac{1}{2}\sum_i a_i^s v_i^s + \frac{1}{2}\sum_i a_i^g v_i^g$ and uses this average instead of the source attention output in the next layer (Gehring et al., 2017).

## 3.2 Training & Inference

Our model is trained on translation bitext by simulating post-edits. Given a bitext corpus, we first train an initial translation system and we then rely on this system to translate the training corpus. This strategy results in three sentences for each example: the source, the guess (i.e. the sentence decoded from the initial system) and the reference sentence. Post-edits are simulated by marking guess tokens which do not appear in the corresponding reference sentence.

The dual attention model presented in the above section is then trained. We maximize the log-likelihood of the training reference sentences $y$ given each corresponding source sentence $x$ and the annotated guess $g$, i.e. we maximize

$$\mathcal{L}_{\text{Train}} : \theta \to \sum_{(x,y,g) \in \text{Train}} \log P(y|x, g, \theta)$$

where $y$ refers to the reference sentence, $x$ refers to the source sentence and $g$ is the annotated guess sentence as defined above. Training relies on stochastic gradient descent (Bottou, 1991), using Nesterov's accelerated gradient with momentum (Nesterov, 1983; Sutskever et al., 2013). At inference time, we decode through standard left-to-right beam search (Sutskever et al., 2014). Our decoding strategy for QuickEdit also incorporates hard constraints that prevent the decoder from outputting tokens which are marked in the guess.

## 3.3 Extension to Monolingual Editing

The extension of QuickEdit to a monolingual setting is straightforward: we remove the source encoder and the corresponding attention path. This results in a single encoder model which takes only an annotated guess as input. This model can be trained from pairs of sentences consisting of a machine translation output along with the corresponding reference sentence. Although machine translation bitext are used to create this model training data, it operates solely on target language sentences without requiring a source sentence at test time. In our experiments, we train distinct models for the monolingual setting. We do not consider sharing parameters with the translation models at this point.

## 4 Experiments & Results

We evaluate on three translation datasets of increasing size and we report results in both language directions: IWSLT'14 German-English (Cettolo et al., 2014), WMT'14 German-English (Luong et al., 2015), and WMT'14 English-French (Bojar et al., 2014). Our post-editing baseline is our initial neural translation system, complemented with decoding constraints to disallow marked guess words to be considered in the beam. For paraphrasing, we compare our model trained on WMT'14 fr-en to the model of (Mallinson et al., 2017) on the MTC dataset (Huang et al., 2002) following their setup. We relied on WMT'14 fr-en training data motivated by its size[1].

For IWSLT'14 we train on 160K sentence pairs and we validate on a random subset of 7,250 sentence-pairs held-out from the original training corpus. We test on the concatenation of *tst2010, tst2011, tst2012, tst2013, dev2010* and *dev2012* comprising 6,750 sentence pairs. The vocabulary for this dataset is 24k for English and 36k for German. For WMT'14 English to German and German to English, we use the same setup as Luong et al. (2015) which comprises 4.5M sentence pairs for training and we test on newstest2014.[2] We took 45k sentences out of the training set for validation purpose. As vocabulary, we learn a joint source and target byte-pair encoding (BPE) with 44k types from the training set (Sennrich et al., 2016b,a). Note that even when using BPE, we solely rely on full word markers, i.e. all the BPE tokens of a given word carry the same binary indication (to be changed/no preference). For WMT'14 English to French and French to English (Bojar et al., 2014), we also rely on BPE with 44k types. This dataset is larger with 35.4M sentences for training and 26k sentences for validation. We rely on newstest2014 for testing[3].

The model architecture settings are borrowed from (Gehring et al., 2017). For IWSLT'14 de-en and IWSLT'14 en-de, we rely on 4-layer encoders and 3-layer decoders, both with 256 hidden units and kernel width 3. The word embedding for source and target as well as the output matrix have 256 dimensions. For WMT'14 en-de and WMT'14 de-en, both encoders and decoders have 15 layers (9 layers with 512 hidden units, 4 layers with 1,024 units followed by 2 layers with 2,048 units). Input embeddings have 768 dimensions, output embedding have 512. For WMT'14 en-fr and WMT'14 fr-en, both encoders and decoders have 15 layers (6 layers with 512 hidden units, 4 layers with 768 units, 3 layers with 1024 units, followed by two larger layers with 2048 and 4096 units). Similar to the German model, input embeddings have 768 dimensions, output embedding have 512 dimensions. For all datasets, we decode using beam search with a beam of size 5.

---

[1] Posterior to our experiments, (Wieting and Gimpel, 2017) released an even large dataset that might be used in our setting.
[2] http://nlp.stanford.edu/projects/nmt
[3] http://www.statmt.org/wmt14/translation-task.html

### 4.1 Post-editing

Our study is based on simulated post-edits, i.e. simulated token deletion actions. We start from machine translation outputs from an initial system in which we label tokens to change automatically. For initial translation, we rely on the convolutional translation system from (Gehring et al., 2017)[4] learned from the training portion of the dataset. For each system output, any word which does not belong to the reference translation is marked to be changed. We perform this operation for the train, validation and test portion of each dataset. The training and validation portion can be used for learning and developing our post-editing system. The test portion is used for evaluation.

Table 1 reports our result on this task. Our QuickEdit method strongly outperforms the baseline post-editing system. Both systems access the same information, i.e. a list of deleted word types, which constrains the decoding. QuickEdit adds attention over the initial sentence with rejection marks. This has a big impact on BLEU. On the larger WMT'14 en-de benchmark, the advantage is over 5 BLEU point for both directions. We conjecture that the improvement is lower on the smaller IWSLT data due to over-fitting, i.e. the base system is excellent on the training set which reduces the post-editing opportunities on the training data, therefore limiting the amount of supervised data for training our post-editing system. We show examples of post-editing from the test set of WMT-14 de-en in Table 2. These examples show the ability of the model to rephrase sentences avoiding the marked tokens while preserving the source meaning. Similar to our experiments on WMT'14 en-de, QuickEdit also reports large improvement with respect to the baseline model on WMT'14 en-fr, with $+5.6$ points (53.4 vs 47.8).

---

[4] https://github.com/facebookresearch/fairseq-py.

One should note that the simulated edits rely on gold information, i.e. crossed-out words are always absent from the reference. Our aim is to simulate a post-editor which might have a sentence close to the reference in mind. This evaluation method allows to conduct large scale experiments without labeling burden. Conducting an interactive post-editing study requires trained editors and interface consideration beyond the scope of this initial work.

### 4.2 Partial Feedback

So far, our post-editing setting marked all incorrect words in the guess. We now consider a setting where the simulated post-editor performs less work by marking only a subset of these tokens. This is analogous to a hypothetical online translation service which offers a feature enabling the user to mark parts of a translation to be improved. In addition to marking only a subset of the incorrect tokens at inference time, we also train new models for which the training data also only had a subset of incorrect tokens marked. Specifically, we train three models QE25, QE50, QE100 for which either $25\%$, $50\%$ or $100\%$ of incorrect guess tokens were marked.

In this setting, we also compare with the baseline model, i.e. the initial translation system augmented with decoding constraints to avoid marked words. Figure 2 plots BLEU as a function of the number of marked words on the validation set of WMT'14 German to English. This curve is obtained by marking at most $1, 2, \ldots, 8$ words to be changed per sentence, taking into account that the actual number of marked word in a sentence cannot be higher than the number of guess words not present in the reference sentence.

Compared to the baseline, there is a small advantage for QuickEdit for 1-2 marked words and a larger improvement when more words are marked. Unsurprisingly, the model trained with fewer marked words (QE25, QE50) performs better when tested with fewer marked words, while QE100 gives the largest improvement with $4$ or more marked words.

### 4.3 Monolingual Editing

Table 1 also reports monolingual results. In that case, the system is not given the source sentence, only a sentence in the target language along with change markers. Even if the model is not given the source, it manages to generate sentences which are closer to the reference than the initial sentences, as shown by the BLEU improvement. This shows the ability of the model to paraphrase from deletion constraints. Table 3 shows examples of the system in action from the English test set of WMT-14 fr-en. This examples show that the model can provide synonyms, e.g. *essential* → *vital*, or *came after* → *followed*. The model can also replace tenses when appropriate, e.g. *have not waited* → *did not wait*, or *wrote* → *had written*.

|  | IWSLT'14 | | WMT'14 (de) | | WMT'14 (fr) | |
|---|---|---|---|---|---|---|
|  | de→en | en→de | de→en | en→de | fr→en | en→fr |
| initial translation | 27.4 | 24.2 | 29.7 | 25.2 | 37.0 | 40.2 |
| post-edit baseline | 33.0 | 30.2 | 34.6 | 30.7 | 45.4 | 47.8 |
| post-edit QuickEdit | 34.6 | 30.8 | 41.3 | 36.6 | 49.7 | 53.4 |
| monolingual QuickEdit | 29.3 | 26.7 | 39.5 | 34.2 | 47.7 | 51.3 |

Table 1: Editing results (BLEU4) when all incorrect tokens are requested to be changed.

| | |
|---|---|
| source | Schauspieler Orlando Bloom hat sich zur Trennung von seiner Frau, Topmodel Miranda Kerr, geäußert. |
| guess | Actor Orlando Bloom ~~has spoken of the~~ separation ~~of~~ his wife, ~~Topmodel~~ Miranda Kerr. |
| output | Actor Orlando Bloom **spoke about** separation **from** his wife, **Top Model** Miranda Kerr. |
| source | Die heutigen elektronischen Geräte geben im Allgemeinen wesentlich weniger Funkstrahlung ab als frühere Generationen. |
| guess | Today's electronic devices generally ~~give far less~~ radio ~~radiation~~ than previous generations. |
| output | Today's electronic devices generally **emit significantly fewer** radio **frequencies** than previous generations. |
| source | Statt sich von der Zahlungsunfähigkeit der US-Regierung verunsichern zu lassen, konzentrierten sich Investoren auf das, was vermutlich mehr zählt: die Federal Reserve. |
| guess | ~~Instead of~~ being ~~obscured~~ by the ~~US~~ government's ~~inability to pay~~, investors ~~focused~~ on what ~~is~~ probably more ~~important~~: the Federal Reserve. |
| output | **Rather than** being **insane** by the **United States** government's **insolvency**, investors **concentrated** on what probably **counts** more: the Federal Reserve. |
| source | Boeing bestreitet die Zahlen von Airbus zu den Sitzmaßen und sagt, es stehe nicht im Ermessen der Hersteller zu entscheiden, wie Fluggesellschaften die Balance zwischen Flugtarifen und Einrichtung gestalten. |
| guess | Boeing is ~~denying~~ the figures ~~from~~ Airbus to the ~~seats~~ and says that it is not ~~left to the discretion of~~ the manufacturers to decide how airlines ~~are~~ to balance air fares and ~~set~~ up. |
| output | Boeing is **contesting** Airbus**'s seating** figures and says it is not **up to** manufacturers to **determine** how airlines balance fares and **equipment**. |

Table 2: Post-editing examples from WMT'14 en-de. Strike-through text indicates the tokens marked to be changed. Bold text indicates tokens introduced by the model, i.e. tokens not present in the original guess.

### 4.4 Paraphrasing

Although it is not our primary goal, monolingual QuickEdit can also be used for paraphrasing by pairing it with another model to automatically generate change markers. In that case, the generative model of edit markers replaces the human instructions. Basically, given an input sentence $x$, the edit model generate a sequence $c$ of binary variables, which indicates whether each word $x_i$ of $x$ should be edited out ($c_i = 1$) or not ($c_i = 0$). QuickEdit then takes $(x, c)$ and generate a sentence $y$ that paraphrases $x$ following the change markers $c$.

| | | |
|---|---|---|
| input | And while ~~the members of~~ Congress ~~cannot~~ agree on whether to ~~continue~~, several ~~States have~~ not ~~waited~~. |
| output | And while **there is no way for** Congress to agree on whether to **go ahead**, several **states did** not **wait**. |
| input | This is ~~truly essential~~ for our nation. |
| output | This is **really vital** for our nation. |
| input | His case ~~came after~~ that of Corporal Glen Kirkland, who ~~told~~ a parliamentary ~~committee~~ last month that he had been ~~pushed out~~ before ~~being~~ ready because he did not meet the universality of service rule. |
| output | His case **followed** that of Corporal Glen Kirkland, who **said to** a parliamentary **panel** last month that he had been **forced to go** before he **was** ready because he did not meet the rule of universality of service. |
| input | Since the ~~beginning~~ of major ~~fighting~~ in Afghanistan, the army has ~~been struggling~~ to determine what latitude it can ~~grant~~ to injured soldiers who want to ~~remain~~ in the ~~ranks~~, but who are not ~~fit for battle~~. |
| output | Since the **start** of major **battles** in Afghanistan, the army has **had a hard time** to determine what latitude it can **give** to injured soldiers who want to **stay** in the army, but who are not **capable** of **battling**. |
| input | ~~Mr.~~ Snowden ~~wrote~~ in his letter that he had ~~been subjected to~~ a ~~serious~~ and sustained campaign of persecution ~~,~~ ~~which~~ forced him ~~to leave~~ his ~~country~~. |
| output | **Mr** Snowden had **written** in his letter that he had **suffered** a **severe** and sustained campaign of persecution that forced him **out** of his **homeland**. |
| input | Spirit Airlines Inc. ~~applied~~ the first ~~hand baggage charges~~ three years ago, and ~~low-cost~~ Allegiant followed ~~a little~~ later. |
| output | Spirit Airlines Inc. **introduced** the first **hand-luggage charge** three years ago, and the **inexpensive** Allegiant followed **somewhat** later. |
| input | "~~I've never~~ seen ~~such a fluid~~ boarding ~~procedure~~ in my ~~entire~~ career"; he ~~says~~. |
| output | "**I have not** seen **this kind of seamless** boarding in my career"; he **said**. |
| input | ~~As a result ,~~ there ~~will~~ be ~~no more~~ employees in the ~~plant~~. |
| output | **This means that** there **won't** be **any** employees in the **factory**. |
| input | Pierre Beaudoin ~~, President and CEO ,~~ is confident ~~that~~ Bombardier ~~will~~ meet its target ~~of~~ 300 firm ~~orders before~~ the first ~~aircraft enters~~ commercial ~~service~~. |
| output | **Chief Executive Officer** Pierre Beaudoin is confident Bombardier **can** meet its 300 firm **order** target **prior to** the first **airplane entering** commercial **services**. |
| input | ~~Another 35 persons~~ involved in ~~trafficking~~ were ~~sentenced to~~ a total of 153 years~~'~~ ~~imprisonment~~ for ~~drug trafficking~~. |
| output | **Thirty-five other people** involved in **the traffic** were **punished with** a total of 153 years in **prison** for **drug-related offenses**. |

Table 3: Monolingual editing examples from the WMT'14 fr-en test set. Strike-through text indicates the tokens marked to be changed. Bold text indicates tokens introduced by the model, i.e. tokens not present in the original guess.

| | Accuracy | Fluency | Boldness |
|---|---|---|---|
| Source | 100% | 100% | 0% |
| ParaNet | 56% | 37% | 16% |
| QuickEdit | 72% | 53% | 21% |

Table 4: Paraphrasing experiments on the MTC dataset.

We use the monolingual QuickEdit model for English trained on WMT-14 fr-en for our paraphrase experiments. We rely on the simplest possible model to generate change markers: for each word type $w$, we estimate its probability to be edited out $P(c_i = 1|x_i = w)$ on the QuickEdit training data based on relative frequency counts.

| | reference | He said that Sino-Kenyan news agencies had long-term cooperative ties and hoped that the ties could further develop in the new century. |
| --- | --- | --- |
| | human | He said the two News Agencies of China and Kenya have friendly relationship over a long period of time. He hoped that this relation could further develop in the new century. |
| | paranet | He said the two **news outlets** in China and Kenya have **amicably similar relationships to** a long period of time. |
| | QuickEdit | He said that the two **news agencies** of China and Kenya **were friends for** a long period of **time** and hoped that the relationship **would continue** in the new century. |
| | reference | Annan urged sharon to ensure israeli forces will "adopt military tactic and weapons that cause a minimum possible threat to safety of palestinian people and personal properties." |
| | human | Annan called on Sharon to ensure that Israeli security forces " use weapons and fighting methods that will cause minimum threat to the safety and property of the Palestinian civilians. " |
| | paranet | Annan called **for** Sharon to " ensure that Israeli security forces **deploy** weapons and **combat** methods that **endanger** security and the property of Palestinian civilians. " |
| | QuickEdit | Annan **calls** on Sharon to " use weapons and **combat practices** that will **pose a** minimum threat to the safety and property of Palestinian civilians. " |
| | reference | [Shuttleworth']s space travel has drawn great publicity in South Africa and won the honor of being the most important news event since Mandela's release from prison. |
| | human | Shuttleworth's space journey has received enormous attention in South Africa and is praised as the most important news since the release of Nelson Mandela from prison. |
| | paranet | Shuttleworth's journey has received enormous attention in South Africa and is **considered** the most important news since the release of Nelson Mandela. |
| | QuickEdit | **The** Shuttleworth space **trip attracted considerable** attention in South Africa and is **lauded** as the most important news since Nelson Mandela **was released** from **jail**. |

Table 5: Paraphrasing experiments on the MTC dataset. Bold indicates tokens introduced by the the models, i.e. tokens which are not in the human source given as input.

For inference, we simply threshold this probability $P(c_i = 1 | x_i = w) > \tau$ to assign change markers. $\tau$ is selected to control how *bold* paraphrasing should be, i.e. large $\tau$ would yield minor changes, while small $\tau$ would edit the input sentence substantially.

We compare our paraphrasing approach with ParaNet (Mallinson et al., 2017), a paraphrasing neural model based on translation pivoting[5]. We conduct our evaluation on the MTC dataset (Huang et al., 2002) following the setup introduced in the ParaNet paper. This setup consists of 75 human paraphrase pairs (excluding duplicate MTC sentences as well as erroneous paraphrases). The evaluation considers each pair of human paraphrases $(x, y)$. Each paraphrasing model (QuickEdit and ParaNet) generates a paraphrase given $x$. Then human judgments are collected by showing $y$ and three versions of $x$, i.e. the original version $x$, its paraphrase from ParaNet $x^{(p)}$ and its paraphrase from QuickEdit $x^{(q)}$. For each example, the three sentences $x, x^{(p)}, x^{(q)}$ are shuffled and do not carry any information about their origin. The assessor should label whether each version of $x$ is a valid paraphrase of $y$ and should rank them by fluency from 1 most fluent to 3 least fluent.

We can evaluate paraphrasing performance at various levels of boldness which we can control with the parameter $\tau$. Bold paraphrasing means that the model needs to generate sentences which differ more from the input $x$ than conservative paraphrasing. In this work, we performed our evaluation using a level of boldness comparable to ParaNet from (Mallinson et al., 2017). Table 4 reports the results of this experiment. Accu-racy measures the fraction of sentences considered valid paraphrases. Fluency measures the number of cases the paraphrase was considered more fluent or as fluent as the source sentence. Boldness measure the fraction of paraphrase tokens that were not in the source.

---
[5]We are thankful to the authors of ParaNet for sharing their generations for our evaluation.

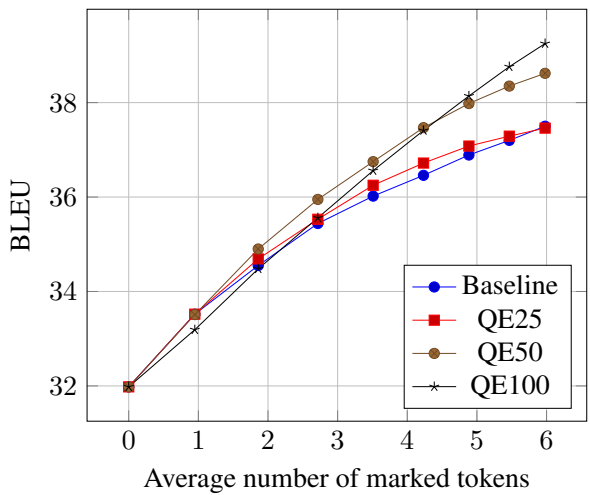

Figure 2: Post-editing results as a function of the average number of marked tokens per sentence on WMT'14 de-en validation set (45k sentences). QE25, QE50, QE100 refer to QuickEdit models trained with data where respectively 25, 50 or 100% of the guess tokens not present in the reference were marked to be changed.

The results highlight the advantages of QuickEdit. The paraphrases from QuickEdit are accurate for 72% of the sentences, as opposed to 56% for ParaNet. The fluency of the generation from QuickEdit ranks equally or higher than the human source sentence for 53% of the examples, which compares to 37% for ParaNet. Table 5 shows a few paraphrases from both models. These examples highlight that the boldness operating point chosen by the authors of ParaNet is rather conservative, with few edits per sentence. Nevertheless, QuickEdit advantage is illustrated in these examples, showing that ParaNet often forgets part of the source sentence while QuickEdit does not, e.g. *could futher develop* in the first example is not expressed by ParaNet but QuickEdit proposes *would continue*. This tendency to shorten the input can yield an opposite meaning, e.g. in the second example, ParaNet rephrases *cause minimum threat* as *endanger* while QuickEdit proposes correctly *pose a minimum threat*. For examples with less conservative paraphrasing, one can refer to Table 3.

## 5 Conclusions

This work proposes QuickEdit, a neural sequence to sequence model that allows one to edit text by simply requesting few initial tokens to be changed. From a marked sentence, the model can generate an edited sentence both in the context of machine translation post-editing (a source sentence is also provided), or in a monolingual setting. In both cases, we assess the impact of the change requests. We show that marking words not present in a hidden reference sentence allow the model to generate text closer to this reference. In the context of post-editing, we conduct simulated post-edits, i.e. we mark words absent from the reference as rejected. We show that crossing out a few words per sentence can drastically improve BLEU, even on top of a strong MT system, e.g. BLEU on WMT'14-en-fr moves from 40.2 to 53.4 with QuickEdit post-editing as opposed to 47.8 for the post-editing baseline. In the context of monolingual editing, we show that our system both allow text editing and paraphrasing. For paraphrasing, we outperform a strong model (Mallinson et al., 2017) in a human evaluation on the MTC dataset, both in terms of accuracy (72% vs 53%) and fluency of the generation (53% vs 37%).

Our work opens several future directions of research. First, we want to extend our evaluation from simulated post-edits to a genuine interactive editing scenario. QuickEdit currently allows only to reject word forms for a whole sentence, not reject them in a specific context. We plan to explore this possibility. Also, QuickEdit could be a good basis for an automatic post-editing system (Chatterjee et al., 2015). QuickEdit can be applied for multi-step editing, letting the user refine their sentence multiple time. In that case, attending to all previous versions of the sentence would be relevant. Finally, we could also consider offering a richer set of simple edit actions. For instance, we could propose span substitutions to the user, which requires a decoding stage proposing a short list of promising spans and candidate replacements.

## Acknowledgments

We thank Marc'Aurelio Ranzato, Sumit Chopra, Roman Novak for helpful discussions. We thank Sergey Edunov, Sam Gross, Myle Ott for writing the fairseq-py toolkit used in our experiments. We thank Jonathan Mallinson, Rico Sennrich, Mirella Lapata, for sharing ParaNet data.